\documentclass{article}
\usepackage{spconf,amsmath,graphicx}
\usepackage[french]{babel}
\usepackage[utf8]{inputenc}
\usepackage{amsfonts}
\usepackage{longtable}
\usepackage{subcaption}
\usepackage{url}

\title{Towards integrating spatial localization in convolutional neural networks for brain image segmentation}

\name{Pierre-Antoine Ganaye, Michaël Sdika, Hugues Benoit-Cattin}
\address{Univ Lyon, INSA‐Lyon, Université Claude Bernard Lyon 1 \\ UJM-Saint Etienne, CNRS, Inserm \\ CREATIS UMR 5220, U1206, F‐69100 \\ Lyon, France}

\begin{document}
%\ninept
%
\maketitle
\begin{abstract}
Semantic segmentation is an established while rapidly evolving field in medical imaging.
In this paper we focus on the segmentation of brain Magnetic Resonance Images (MRI) into cerebral structures using convolutional neural networks (CNN). CNNs achieve good performance by finding effective high dimensional image features describing the patch content only. In this work, we propose different ways to introduce spatial constraints into the network to further reduce prediction inconsistencies.

A patch based CNN architecture was trained, making use of multiple scales to gather contextual information.
Spatial constraints were introduced within the CNN through a distance to landmarks feature or through the integration of a probability atlas.
We demonstrate experimentally that using spatial information helps to reduce segmentation inconsistencies.

\end{abstract}

\begin{figure*}[!h]
  \begin{subfigure}{1\linewidth}
  	\centering
    \includegraphics[width=0.85\linewidth]{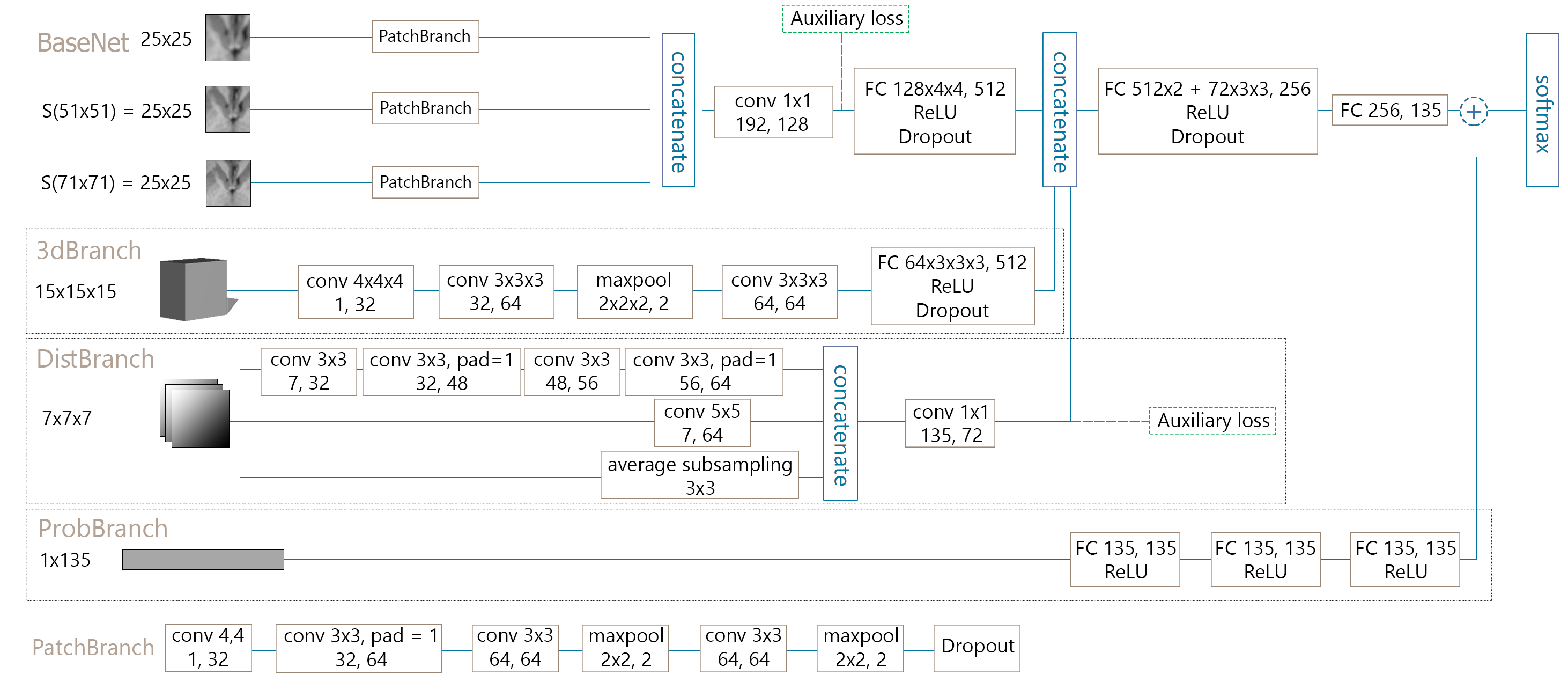}
  \end{subfigure}
  \caption{Architecture of the network, composed of a 2D multi-resolution CNN (BaseNet), 3dBranch for the 3D input, DistBranch for the distances to landmarks and ProbBranch to integrate knowledge from the probability atlas.}
  \label{fig:cnnarch}
\end{figure*}

\section{Introduction}
\label{sec:intro}
Multi-Atlas methods\cite{HECKEMANN2006115, mindboogle} are one of the main approaches for the segmentation of brain structures: the segmentation maps produced individually by a set of atlas are combined for better quality. Machine learning can be used to improve the segmentation mapping or the label fusion of multi-atlas methods \cite{SDIK-15b,wang2013multi}. It can also be used by itself for MR brain image segmentation \cite{Stollenga:2015:PML:2969442.2969574,DBLP:journals/corr/RoyCSKNW17,moeskops_automatic_2016}. \newline

Following the democratization of deep neural networks, \cite{lee} proposed an approach where a CNN is trained to predict the class of the center pixel of a patch. The segmentation map is obtained by applying the trained model as a sliding window over the image. Other variants \cite{moeskops_automatic_2016, brebisson} integrate patches of different sizes and resolutions, across parallel branches, in order to bring more contextual information to the network.

Recently another contribution based on an encoder-decoder architecture, shown its efficiency for brain segmentation \cite{DBLP:journals/corr/RoyCSKNW17} by firstly pre-training a CNN on a dataset annotated with FreeSurfer and optimizing a custom loss function during fine-tuning. This method produced state of the art segmentation maps. However, this type of architecture requires storing large features maps, which leads to memory consumptions issues. In this paper we chose to work on patch-based segmentation models, which provide interesting modeling properties to partially constrained problems.

Multi-atlas segmentation approaches \cite{HECKEMANN2006115, mindboogle, SDIK-15b, wang2013multi} based on diffeomorphic registration methods are known to preserve the topology of the structures.
Unlike atlas based segmentation methods, patch based CNN do not use the spatial position of the patch within the image volume.
We propose and evaluate several ways to incorporate this prior knowledge to existing by-patch architectures, in order to yield better constraints learned from spatial features.

The integration of spatial information in a brain segmentation model was explored in \cite{ANBEEK2005795}, where a nearest neighbor classifier combines voxel level intensities and spatial knowledge.
Patch-based segmentation CNN using spatial features have also been proposed in the literature, \cite{ghafoorian_deep_2017} used a combination of spatial coordinates and landmarks, which requires a pre-segmentation, \cite{brebisson} proposed to use a vector composed of distance to centroids, which requires to iterate a number of times to improve the estimates.
In this paper, unlike \cite{brebisson} and \cite{ghafoorian_deep_2017}, we propose an architecture to integrate spatial context knowledge in any patch based CNN segmentation method, without requiring any kind of initialization nor iteration and is robust to degenerate cases such as nested structures. \newline

We show the benefit of our approach, which does not require additional annotation, by applying it to a multi-resolution CNN, on the multi-atlas brain segmentation challenge of MICCAI 2012.

In section 2. we first present the multi-resolution CNN and show how we introduce spatial context and information a priori. In section 3. we present the dataset and the evaluation metrics, followed by the implementation details. Finally in section 4. we illustrate and comment the obtained results.

\section{Methods}
Our segmentation model is composed of a multi-resolution CNN (BaseNet), which is extended by tree additional branches (3dBranch, DistBranch, ProbBranch), each bringing more information and spatial context. 
Note that each of the additional branch can be used separately, as an individual component.

\subsection{Multi-Resolution CNN}
In Fig \ref{fig:cnnarch}, we describe the main 2D multi-resolution classifier, named BaseNet in the remainder of the paper. It is composed of 3 branches inspired from \cite{moeskops_automatic_2016, brebisson}.
The three branches are similar, having as input 25x25 patches but with three different scales. The two coarser patches are created by subsampling the original ones (of size $51^2$ and $71^2$), so that a larger context is accounted without increasing the number of parameters.
The features produced by the three branches are concatenated and a 1x1 convolution is applied in order to reduce the feature space while keeping consistent information. Finally three consecutive fully connected layers are used to produce the scores together with the softmax function. To take advantage of the volumetric nature of the data, we try to test the importance of using a 3D branch (3dBranch). For doing this, a patch of size $15^3$ is extracted and merged into the network with a dedicated branch. 
The size of the 3D patch is chosen to limit the brutal increase of parameters due to convolutions with 3D kernels while focusing on the fine scale information. \newline
The BaseNet architecture is used as an initial basis block throughout this article and will serve as reference to measure the performance of the proposed following branches.

\subsection{Spatial information}
Although the spatial position within the brain is a relevant information for brain structure segmentation, the classification produced by the BaseNet CNN relies on the patch content only. We propose to introduce a spatial representation of the patch position.

Let $\mathcal{L}$ be a set of landmarks defined uniformly along each axis of the input volume, where $\mathcal{L} \subset \mathbb{R}^{3}$. For a patch centered at position $x$, we evaluate $D \in \mathbb{R}^{|\mathcal{L}|}$, the euclidean distance vector of $x$ to each landmarks in $\mathcal{L}$.
Unlike \cite{brebisson,ghafoorian_deep_2017}, our landmarks are evenly distributed along each axis of the input image, they are not related to any region of the brain, thus do not require any pre-segmentation.
As the landmarks are on a regular grid, $D$ can be represented as a 3D image, which enables the use of convolutional layers.
This distance image $D$ serves as input to the DistBranch composed of convolutions with different kernel sizes, inspired from the inception module in \cite{szegedy2017inception} and finally merged with the second fully connected layer.
The use of a radial basis function kernel to normalize the input distance image $D$ has also been evaluated:
$$rbf(D) = \exp(-\alpha D^2)   \;\; \alpha \in \mathbb{R}^+$$

\subsection{A priori via probability atlas}
We also considered to account for the voxel position into the BaseNet segmentation by merging the CNN probability output with a more standard single probabilistic atlas segmentation. For a given voxel, the class probability vector according to the probabilistic atlas passes through three fully connected layers of size $\ell$ each, where $\ell$ is the number of classes. The output is summed to the BaseNet output just before the softmax function. This change is described as ProbBranch in Fig \ref{fig:cnnarch}.

\begin{table*}[t]
  \centering
  \begin{tabular}{|l|c|c|c|c|}
    \hline
    Model & Dice & Hausdorff & MSD & $N_{\text{param}}$ \\
    \hline
    BaseNet & 0.694 $\pm$ 0.17 & 40.26 $\pm$ 40.12 & 1.74 $\pm$ 2.14 & 1 249 415\\
    BaseNet + DistBranch & 0.720 $\pm$ 0.14 &  10.09 $\pm$ 5.41 & 1.10 $\pm$ 0.64 & 1 508 511\\
    BaseNet + ProbBranch & 0.700 $\pm$ 0.17 & 32.38 $\pm$ 36.90 & 1.50 $\pm$ 1.80 & 1 304 090\\
    BaseNet + DistBranch + ProbBranch& 0.723 $\pm$ 0.14 & 9.95 $\pm$ 5.29  & 1.10 $\pm$ 0.65 & 1 563 186\\
    BaseNet + DistBranch + ProbBranch + 3dBranch& 0.733 $\pm$ 0.14 & 9.99 $\pm$ 5.63 & 1.07 $\pm$ 0.63 & 2 847 794\\
    Full (all branches + augmentation + supervision) & 0.748 $\pm$ 0.14 & 9.66 $\pm$ 5.46 & 1.00 $\pm$ 0.59 & 2 847 794\\
    \hline
    UNet (with max unpooling) \cite{DBLP:journals/corr/RoyCSKNW17} & 0.708 $\pm$ 0.16 & 51.92 $\pm$ 40.73 & 2.14 $\pm$ 3.01 & 599 040\\
    \hline
  \end{tabular}
  \caption{Distance and similarity metrics for each best performing models. MSD is the mean surface distance and $N_{\text{param}}$ is the number of parameters of the network. All the metrics are averaged over the test dataset. (average $\pm$ standard deviation)}
  \label{tab:results}
\end{table*}

\section{Experiments}
\subsection{Data}
The dataset is composed of 1.5T MRI from the OASIS project, it was distributed during the multi-atlas segmentation challenge of MICCAI 2012. The images were manually segmented into $\ell=135$ classes (structures and background). The original training dataset (15 images) was split into two non-overlapping sets : training (10 images), validation (5 images). The test dataset (20 images) is used to assert the performance of the models on unseen data. The images were affine registered to a reference atlas with FSL Flirt\cite{jenkinson2002}.
All the images were skull striped, using a brain extraction software. The mean and standard deviation were estimated on the training set, all the images were finally mean centered and reduced.

\subsection{Evaluation}
The different methods discussed in this work were evaluated using the Dice coefficient,
the Hausdorff distance $d_H$ and the mean surface distance $d_{msd}$:
$$dice(X,Y) = \frac{2|X \cap Y|}{|X| + |Y|},$$
%$$d_H(X,Y) = max \{ \sup_{x \in \mathit{X}} \inf_{y \in \mathit{Y}} d(x,y), \sup_{y \in \mathit{Y}} \inf_{x \in \mathit{X}} d(x,y) \},$$
%%$$d_{msd}(X,Y) = \frac{1}{2}\left(\displaystyle\sum_{x \in \mathit{X}} \frac{\inf_{y \in \mathit{Y}} d(x,y)}{|\mathit{X}|} + \displaystyle\sum_{y \in \mathit{Y}} \frac{\inf_{x \in \mathit{X}} d(x,y)}{|\mathit{Y}|}\right),$$
%$$d_{msd}(X,Y) = \frac{1}{2|\mathit{X}|}\displaystyle\sum_{x \in \mathit{X}} \inf_{y \in \mathit{Y}} d(x,y) + \frac{1}{2|\mathit{Y}|} \displaystyle\sum_{y \in \mathit{Y}} \inf_{x \in \mathit{X}} d(x,y),$$
$$d_H(X,Y) = \max \{ \sup_{x \in \mathit{X}} d(x,Y), \sup_{y \in \mathit{Y}} d(y,X) \},$$
$$d_{msd}(X,Y) = \frac{1}{2}\left(\displaystyle\sum_{x \in \mathit{X}} \frac{d(x,Y)}{|\mathit{X}|} + \displaystyle\sum_{y \in \mathit{Y}} \frac{d(y,X)}{|\mathit{Y}|}\right),$$
where $\mathit{X}$ and $\mathit{Y}$ are two segmentation maps of a same label. $d$ is defined as the minimal distance between a point and a set.

\subsection{Implementation details}
The number of landmarks was fine-tuned experimentally on the validation dataset, by varying the parameter from $3^3$ to $10^3$ points. We noticed an increase in precision until $7^3$ and kept this parameter as a good balance between performance and processing time.
%To integrate the distances during the training of the network, we tried to consider the data as a 1D signal, 2D image and 3D volume. 
We evaluated several representations for the distance image $D$ : a 1D vector, a set of 2D images or a 3D volume.
The 2D approach showed to be a good balance between the moderate performance of the 1D model and the cost of the 3D model because of the 3D convolutions. In the radial basis function, the value of $\alpha$ was set to $0.01$.

The cross-entropy is used as the cost function. The numerical optimization was performed with SGD, with an initial learning rate $lr_0 = 1e-3$ and a momentum of 0.9. As in \cite{DBLP:journals/corr/ChenPK0Y16} the learning rate was updated at each epoch with the poly rate policy:
$$poly(iter) = lr_{0} *\left(1-\frac{iter}{max_{iter}}\right)^{power}$$
Where $iter$ is the index of the epoch, $max_{iter}$ the maximum number of epochs and $power=0.9$. Batch of size 256 gave the best results.
ReLU is the default activation function. Our 'Full' model is composed of all the branches, uses auxiliary losses and data augmentation.

\subsubsection{Regularization \& Auxiliary Loss}
To prevent overfitting of the model, we used $l_2$ regularization, defined by an additional penalty term on the objective function $\frac{\lambda}{2} w^2$.
%, where $w$ are the weights and $\lambda$ is the regularization parameter controlling the trade off. 
Dropout \cite{srivastava2014dropout} was used for regularization, by randomly setting units output to 0.
To ease the training of the network, we used auxiliary cost functions as advised in \cite{szegedy2017inception}, at two different locations (BaseNet, DistBranch). An auxiliary loss consists of a fully connected layer attached to a branch, followed by a softmax function. The global loss is composed by the network's loss and the auxiliary losses.

\subsubsection{Class imbalance \& Data Augmentation}
Because the anatomical regions of the brain have varying volumes, sampling from the original distribution produces class imbalance. We tried to balance the classes by adjusting the weights accordingly in the cross-entropy cost function, unfortunately we did not see any improvement. \newline
In order to increase the variability of the images for better generalization, random data augmentations were applied on the largest 2D patch, by combining rescale with a factor in the range [0.9;1.1] and rotation in the range [-10;10] degrees.

%This model was implemented in Python using the PyTorch framework \cite{pytorch}. This work also relies on open-source libraries such as Scipy, SimpleITK and NiBabel.

\begin{figure*}[t!]
    \centering
        \includegraphics[height=1.35in]{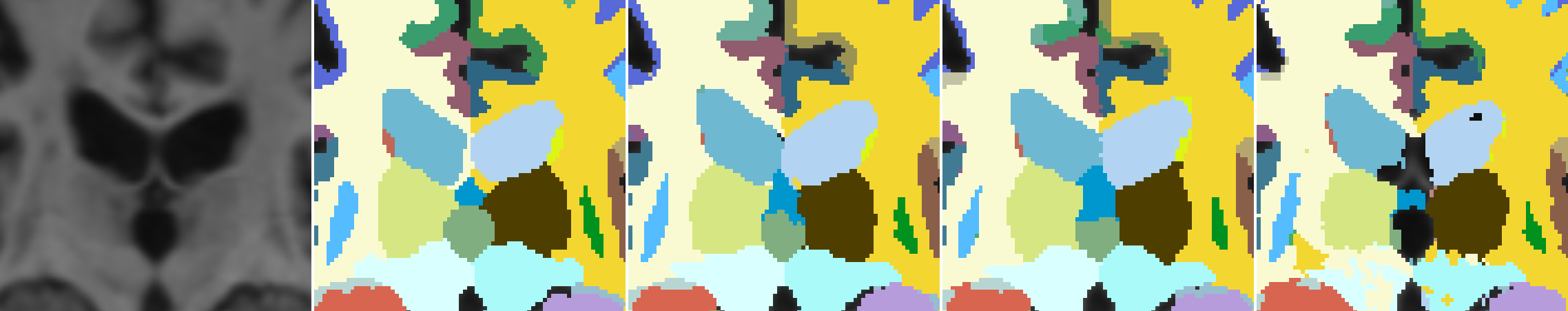} \\
                (a) \hspace*{30mm} (b) \hspace*{30mm} (c) \hspace*{30mm} (d) \hspace*{30mm} (e)
    \caption{Illustration of the segmentations. Coronal slice (a) and associated segmentation maps : ground truth (b), Full (c), BaseNet+DistBranch (d) and BaseNet (e). The segmentations were obtained on one patient of the test dataset.}
    \label{segmaps}
\end{figure*}

\newpage
\section{Results}
To evaluate the optimal way of integrating the distances to landmarks into the BaseNet CNN, we tested the performance of four architectures, table \ref{tab:distance} shows the results obtained. It is clear that using the radial basis function to normalize does not really help, consequently we chose not to include it. As a matter of fact, it seems the CNN is able to adjust the signal values with the help of linear operations, such as convolutions. This assumption is underpinned by the poor result of the model 'distance with rbf'. We decide ultimately to integrate 2D distances into the network with convolution layers and without any kind of normalization.

\begin{table}
  \centering
  \begin{tabular}{l|c|c}
    Model & Dice & Hausdorff \\
    \hline
    distance & 0.703 $\pm$ 0.17 & 15.84 $\pm$ 12.04\\
    distance + rbf & 0.452 $\pm$ 0.36 & 66.02 $\pm$ 40.19\\
    distance + conv & 0.720 $\pm$ 0.14 & 10.09 $\pm$ 5.41\\
    distance + conv + rbf & 0.718 $\pm$ 0.17 & 10.42 $\pm$ 5.82\\
  \end{tabular}
  \caption{Comparison of methods to integrate distance to landmarks, performance measured using the dice similarity and the Hausdorff distance. (average $\pm$ standard deviation)}
  \label{tab:distance}
\end{table}
Figure \ref{segmaps} shows an example of segmentation maps we produced with the tested models. A real performance gap can be noticed between BaseNet(e) and BaseNet+DistBranch (d), where the first detects background between the left and right lateral ventricles and the second is able to recover smooth structures.

In table \ref{tab:results}, we can notice the impact of each branch in this incremental setup. Adding each of them successively brings better results. 
%The best trades between increase in performance and model complexity are obtained with the distance to landmarks branch (DistBranch) and the data augmentation combined with auxiliary losses. 
The 2D multi-resolution model (BaseNet) combined with the distance integration (DistBranch) shows a noticeable decrease in the average and standard deviation of the Hausdorff distance, thus reducing serious segmentation issues, with the help of better spatial constraints. The best model is finally a combination of all the proposed branches, leading to an average dice of $0.748$. In comparison, \cite{brebisson} which is the only to our knowledge to have used a by-patch segmentation approach for the original 135 classes problem, proposed a model composed of 30M parameters and reached an average dice of $0.725$. Our model has 10 order of magnitude less parameters, with a better average dice. With this model, we would have been ranked 5th of the multi-atlas segmentation challenge at MICCAI 2012, with a segmentation time per image of approximately 9 minutes. 

We briefly compare to a UNet \cite{unet} like encoder-decoder architecture inspired from \cite{DBLP:journals/corr/RoyCSKNW17}, with skip-connections and max unpooling. It was trained to segment slice by slice, optimized only with cross-entropy and dice loss, on the same dataset. It showed encouraging dice similarity, but poor Hausdorff performance, demonstrating that patch based segmentation is still a competitive task for brain segmentation.

\section{Conclusion}
In this study, we set a framework to integrate spatial constraints into any patch based classification network. We showed that integrating distances to landmarks into a 2D multi-resolution CNN can help reduce segmentation incoherence. Adding information from a probability atlas together with 3D patch helps to minimize segmentation errors further. \newline
Encoder-decoder segmentation architecture showed promising results in terms of speed and performance, future work could be done to introduce specific spatial constraints in such model to reduce segmentation inconsistencies.
% $$||x - l_{ijh}||^{2}$$
% $$||T(x) - l_{ijh}||^{2}$$

\section{Acknowledgement}
This work was funded by the CNRS PEPS "ReaDAPTIM" and was performed within the framework of the LABEX PRIMES (ANR-11-LABX-0063) of Université de Lyon, within the program ”Investissements d’Avenir”(ANR-11-IDEX-0007) operated by the French National Research Agency (ANR). We would like to thank the E.E.A doctoral school for providing a financial support.
We gratefully acknowledge the support of NVIDIA Corporation with the donation of the Titan X Pascal GPU used for this research. Also we would like to thank the IN2P3 computing center for sharing their resources.


\bibliographystyle{IEEEbib}

\end{document}